\documentclass{article} 
\usepackage{iclr2021_conference,times}
\usepackage{graphicx}

\usepackage{amsmath,amsfonts,bm}

\def\eqref#1{equation~\ref{#1}}

\def\1{\bm{1}}

\DeclareMathAlphabet{\mathsfit}{\encodingdefault}{\sfdefault}{m}{sl}
\SetMathAlphabet{\mathsfit}{bold}{\encodingdefault}{\sfdefault}{bx}{n}

\newcommand{\E}{\mathbb{E}}

\newcommand{\R}{\mathbb{R}}

\usepackage{float}
\usepackage{hyperref}
\usepackage{url}
\usepackage{xurl}

\title{Beyond Labels: Visual Representations for Bone Marrow Cell Morphology Recognition}

\iclrfinalcopy
\author{Shayan Fazeli \& Majid Sarrafzadeh \\
Department of Computer Science\\
University of California, Los Angeles\\
Los Angeles, California, USA, 90095 \\
\texttt{\{shayan,majid\}@cs.ucla.edu} \\
\And
Alireza Samiei \& Thomas D. Lee \\
Department of Pathology and Laboratory Medicine\\ David Geffen School of Medicine\\ University of California, Los Angeles \\Los Angeles, California, USA, 90095\\
\texttt{\{asamiei,tdlee\}@mednet.ucla.edu} \\
}

\begin{document}

\maketitle

\begin{abstract}
Analyzing and inspecting bone marrow cell cytomorphology is a critical but highly complex and time-consuming component of hematopathology diagnosis. Recent advancements in artificial intelligence have paved the way for the application of deep learning algorithms to complex medical tasks. Nevertheless, there are many challenges in applying effective learning algorithms to medical image analysis, such as the lack of sufficient and reliably annotated training datasets and the highly class-imbalanced nature of most medical data. Here, we improve on the state-of-the-art methodologies of bone marrow cell recognition by deviating from sole reliance on labeled data and leveraging self-supervision in training our learning models. We investigate our approach's effectiveness in identifying bone marrow cell types. Our experiments demonstrate significant performance improvements in conducting different bone marrow cell recognition tasks compared to the current state-of-the-art methodologies. Our codes and toolkits are available at \href{https://github.com/shayanfazeli/marrovision}{https://github.com/shayanfazeli/marrovision}.
\end{abstract}

\section{Introduction}
Bone marrow biopsy is a crucial component of hematopathology. 
It is used to diagnose neoplastic and non-neoplastic blood disorders. 
As part of the biopsy, an aspirate slide is created by smearing the bone marrow spicules on a glass slide to spread the cells and allow an expert hematopathologist to perform a morphological examination of individual cells. 
Specifically, a hematopathologist assesses the cells for possible morphological aberrancies and generates a differential cell count of various cell types. 
According to the International Council for Standardization in Hematology (ICSH) and the World Health Organization (WHO), the differential cell count is a required component of the diagnostic criteria for many neoplastic and non-neoplastic disorders, such as acute leukemias, myelodysplastic syndromes, and plasma cell dyscrasias \citep{lee2008icsh,swerdlow2008classification}. 

The ICSH recommends that when a precise percentage of an abnormal cell type is required for diagnosis, at least 500 bone marrow cells are to be counted in a minimum of two bone marrow smears to generate a representative percentage of individual cell lines \citep{lee2008icsh}. Nonetheless, generating differential cell counts on individual bone marrow biopsies is a time-consuming task for busy hematopathology services. 
Given the prevalence and necessity of bone-marrow biopsy, which is carried out over half a million times in the United States every year, proposing automated methodologies can significantly reduce the costs associated with the laborious and error-prone process of manually reviewing the images
\citep{kennethsymington_2022}.

Recent machine learning advances in the computer vision domain have allowed researchers to apply deep neural inference pipelines to automate medical image analysis. Nevertheless, there are major challenges associated with applying these pipelines in the medical image analysis field. 

First, effective training of deep neural networks often requires access to large, representative datasets with millions of images. However, healthcare datasets, which are commonly generated at large academic hospitals, reflect the patient volume and population-specific to that institution. These datasets are small, in the order of thousands, and are biased towards a specific population of patients who are referred to tertiary academic centers.
 
 \begin{figure}[h]
    \centering
    \includegraphics[width=1\textwidth]{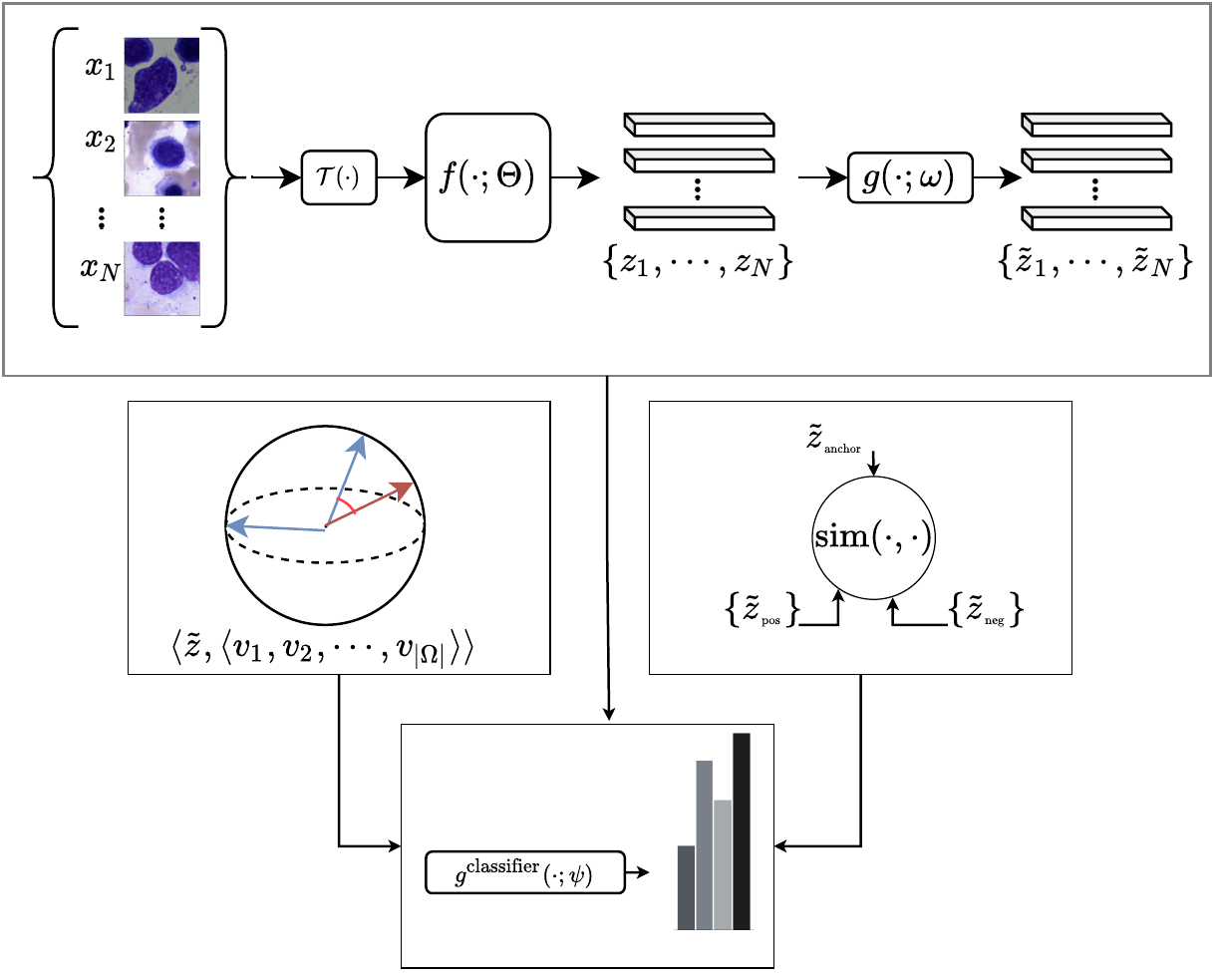}
    \caption{
Pipeline overview - The images of cells are first processed through a deep network. The latent representations then undergo the required projections and inter-item similarity assessments, and the losses for regularizing model weights are computed. Lastly, the latent representation $z$ will be used to determine the cell type in the final classification layer.
    }
    \label{fig:overview}
\end{figure}

Second, expert-annotated datasets are scarce in healthcare. Annotating medical images is labor-intensive, time-consuming, and in some cases, noisy due to inter-annotator disagreement. Many medical images which are prone to inter-annotator variabilities, such as HER2 score staining in pathology, typically require three independent expert annotations for reliable ground truth labeling. Consequently, it is more expensive and laborious to generate these datasets. 

Third, deep learning algorithms' performance peaks on balanced datasets in which there is little variability between the number of samples in each class. However, medical datasets are highly imbalanced and exhibit long-tailed distributions in which rare diseases or cell types (tail classes) are under-represented compared to common diseases or cell types (termed head classes). This implicit bias in data representation can lead deep learning models to excel on over-represented samples in the dataset while potentially neglecting and under-performing on the less common samples.

These challenges have prompted us to focus on leveraging images' intrinsic features to label and classify medical images.

Here, we propose self-supervision as a powerful tool to increase the efficacy and generalizability of visual representations for bone marrow cell morphology classification. Our contributions are as follows:

\begin{itemize}
    \item{
        We have reproduced and improved upon the state-of-the-art supervised model of bone marrow cells' cytomorphologic classification by providing an improved supervised baseline.
    }
    \item{
        We propose a novel solution to the bone-marrow cell cytomorphology classification task by leveraging the recent advancements in self-supervised learning. 
        We have investigated the effectiveness of invariance-centered self-supervision techniques in polishing the learned representations in the latent space.
        We provide empirical findings emphasizing the importance of our solution in generating separable and interpretable features.
        This is shown via training a shallow linear classifier leveraging fixed versions of our latent representations and evaluating the resulting network on a bone marrow cell cytomorphology benchmark.
    }
    \item{
        We investigate the impact of leveraging both supervised signal from the expert annotations as well as the contrast and invariance regularization by optimizing Supervised Contrastive loss.
    }
    \item{
    Based on our findings, we propose that increased reliance on self-supervised learning has the potential to significantly improve performance in medical image classification domains, such as bone marrow cell classification. 
    }
    \item{
    We demonstrate that self-supervised learning increases prediction fairness by rendering visual representations more robust to the negative impacts of long-tail label distributions. Therefore, self-supervised learning is particularly effective in improving classification performance on under-represented classes.
    }
    \item{
    We have made our codes and toolkits publicly available to accelerate the introduction and utilization of self-supervised learning algorithms in medicine. 
    }
\end{itemize}

\section{Related Works}
The rise of convolutional neural networks (CNNs) in 2012 opened the door for the efficient training of deep inference pipelines on large-scale image recognition datasets \citep{krizhevsky2012imagenet}.
Since then, the performance of CNNs has been significantly improved by the introduction of advanced pipelines such as ResNet and ResNeXt architectures \citep{he2016deep,xie2017aggregated}.
To this day, most learning architectures are built on supervised learning algorithms, in which annotated data is required to train the network. However, there have been significant advancements in unsupervised learning algorithms and image representation techniques in the recent computer vision literature. Specifically, significant progress has been made in extracting optimal input representation using auto-encoders and contrastive learning, in which the network extracts and learns image representations invariant to various transformations of observations sharing the same identity, such as rotated or altered context views \citep{chen2020simple,chen2020big,caron2020unsupervised,grill2020bootstrap,li2022neural,yu2020learning,pu2016variational}.

Many of these supervised and unsupervised architectures have been applied in the medical field, particularly to image processing tasks in radiology and pathology
\citep{johnson2019mimic,johnson2019mimic1,mcfarland2018improved}.
In comparison, the effectiveness of deep learning in histology and histopathology is less investigated. 
\citet{matek2021highly} have recently released a publicly available comprehensive bone marrow cell cytomorphology corpus. The authors have also established a highly accurate benchmark for bone marrow cell classification using a supervised deep convolutional neural network.


Here, we present our approach, which further utilizes the semantic context of the cell images in training via self-supervised techniques and improves the current state-of-the-art performance for determining the type of bone marrow cells.

\section{Methods}

\subsection{Supervised}
We prepared an efficient self-supervised learning inference pipeline to improve the current state-of-the-art and set a new self-supervised learning baseline for the bone marrow cell cytomorphology task.
Similar to \citet{matek2021highly}, we have ResNeXt-50 at the core of the supervised pipeline. 
The pipeline in \citet{matek2021highly} employs specific color perturbations and augmentations, adopted from \citet{tellez2019quantifying}. 
We did not implement these augmentation strategies to limit the confounding effect of such pre/post-processing modules while inspecting our model's effectiveness. 
We propose that a stable training procedure with a longer epoch sequence can further improve performance compared to task-specific histopathology pre-processing transformations. 

Additionally, \citet{matek2021highly} 's dataset class distribution is highly imbalanced, reflecting the actual imbalanced distribution of various cell types in the human bone marrow tissue. 
As such, we used balanced sampling and stratified splitting in our pipeline to deal with this considerable class imbalance. 
This was particularly beneficial in preventing the under-penalization of images sampled from less frequent categories. 
In addition, we prepared a supervised baseline performance following the {\it mixup} augmentation protocol to further improve the supervised baseline performance and to smoothen the decision boundaries \citep{zhang2017mixup}.

The assumptions and setup for our supervised inference pipeline are as follows: 
Let $\mathcal{D}$ represent \citet{matek2021highly} dataset's expert image annotations, in which $\mathcal{D}=\{(x_i,y_i)\}_{i=1}^{|\mathcal{D}|}$.
Each $x_i$ is an RGB image of dimensions $H \times W \times 3$ corresponding to a single bone marrow image from the dataset, with the assumption that magnification and scale are consistent throughout the dataset. $\mathcal{T}(\cdot)$ is a probabilistic transformation including standardization and flips along horizontal and vertical axes. Given each $(x_i, y_i)$ sample, class probabilities for all $j \in \{1, 2, \cdots, C\}$ categories are computed. First, the image is mapped to the semantic space through $f(\cdot ; \Theta)$:

\begin{equation}
\label{latent_rep}
z_i = f(\mathcal{T}(x_i) ; \Theta) \; \; \; \; i \in [N]=\{1, 2, \cdots, N\} \; \; \text{(Latent Representation)}
\end{equation}

Having the latent representation $z \in \mathbb{R}^{d_{\text{emb}}}$, class probabilities are computed by performing softmax operation on the dot product of the latent representation $z_i$ and class representations ($w_j$ is the $j$th column of $W_{\text{cls}}$, where  $W_{\text{cls}}$ and $b_\text{cls}$ are weights and biases of the last projection layer.
\begin{equation}
p_i^{(j)} = \cfrac{
\exp(w_j^T \cdot z_i + b_j)
}{
\sum_{k=1}^{C} \exp(w_k^T \cdot z_i + b_j)
}
\end{equation}
$p_i^{(j)}$ gives us the empirical approximation of $\Pr\{Y_i = j | X_i=x\}$. The objective is then to minimize the following cross-entropy loss:

\begin{equation}
    l_i = - \sum_{j=1}^{C} y_i^{(j)} \cdot \ln p_i^{(j)} 
\end{equation}
\begin{equation}
\label{ce_loss}
    \mathcal{L} = \cfrac{1}{N}\sum_{i=1}^N l_i = -\cfrac{1}{N}\sum_{i=1}^N \sum_{j=1}^{C} y_i^{(j)} \cdot \ln p_i^{(j)} 
\end{equation}

One of the known disadvantages of cross-entropy loss is 
producing uncalibrated probability distributions, leading to high-confidence misclassification of rare classes. We use weighted cross-entropy and balanced sampling to alleviate such effects.
Balanced sampling ensures that in each epoch, each example is seen $N_c$ times. Hence, in Equation \ref{ce_loss} we have $N = C \cdot N_c$.

\subsection{Self-supervised}

We trained a visual recognition pipeline following the SwAV strategy to learn visual representations without using any labels or expert annotations \citep{caron2020unsupervised}. We use ResNet-50 architecture at the core of the model. We demonstrate that upon convergence, it leads to performance comparable to the supervised scheme and even surpasses its performance in most tasks.

Having the dataset $\mathcal{D}=\{(x_i, y_i)\}_{i=1}^{|\mathcal{D}|}$, we ignored all of the labels for pretraining and considered $\mathcal{D}^{\text{unsupervised}}$, which only contains  the datapoints. 

Since we are not using cross-entropy, the implicit bias due to label imbalance is less impactful on the final training outcome. As we are working with the imbalanced class distribution of the original dataset, we have $N = |\mathcal{D}|$. 


The latent representations are computed using multiple transformed versions of each image according to Equation \ref{latent_rep}. These representations are then fed to a projection head $g(\cdot\; \omega)$. The resulting projected representations of $m$th view of each example $x_i$ is $u^{(m)}_i \in \mathbb{R}^d_{\text{proj}}$:
\begin{equation}
    u^{(m), \text{raw}}_i=g(z^{(m)}_i ; \omega) \;  \; m \in \{1,2\} \; \wedge \; i \in [N]=\{1, 2, \cdots, N\} \; \; \text{(Projected Representations)}
\end{equation}

\begin{equation}
    u^{(m)}_i=\cfrac{u^{(m), \text{raw}}_i}{||u^{(m), \text{raw}}_i||_2} \; \; \; \text{(L2 Normalization)}
\end{equation}

A prototype set of $\Omega = \langle v_1, v_2, \cdots, v_{|\Omega|} \rangle$ is prepared and initialized uniformly on the $d_{\text{proj}}$-dimensional unit ball to effectively capture the common attributes of images. For each example, the similarities of the corresponding projection and prototype vectors give a predicted {\it code} vector $c \in \mathbb{R}^{|\Omega|}$ such that:

\begin{equation}
    c^{(m)}_i[j] =  \text{sim}(u^{(m)}_i, v_j)
\end{equation}

The prototypes are also assigned to the examples according to the optimal control protocol and via the Sinkhorn algorithm, leading to pseudo-groundtruth distributions $o_i \in \mathbb{R}^{|\Omega|}$. The SwAv loss is then computed as below:

\begin{equation}
    l_i = \cfrac{-1}{M(M-1)} \sum_{m_1=1}^M \sum_{\substack{m_2=1\\ m_2 \not = m_1}}^M \sum_{j=1}^{|\Omega|} o^{(m_1)}_i[j] \ln c^{(m_2)}_i[j]
\end{equation}

\begin{equation}
    \mathcal{L} = \cfrac{1}{N} \sum_{i=1}^N l_i = \cfrac{-1}{NM(M-1)} \sum_{i=1}^N \sum_{m_1=1}^M \sum_{\substack{m_2=1\\ m_2 \not = m_1}}^M \sum_{j=1}^{|\Omega|} o^{(m_1)}_i[j] \ln c^{(m_2)}_i[j]
\end{equation}

Intuitively, this loss encourages the examples that are {\it similar but distinct}, meaning that they share the same identity, such as different views of a single image, to predict each others' codes. Given that the batch size $B$ is relatively small for the Sinkhorn algorithm to work effectively, we also have the queue $Q$ to facilitate better assignments. Once the training is done, the core CNN is frozen, and a linear projection for classification with parameters $\langle W_{\text{cls}}, b_{\text{cls}} \rangle$ are introduced and trained by optimizing the cross-entropy loss (Equation \ref{ce_loss}) to perform the task by leveraging these semi-supervised representations.

\subsection{Supervised Contrastive}
Cross-entropy loss is the de-facto for training deep inference pipelines. 
Nevertheless, it suffers from challenges leading to an increased likelihood of non-calibrated predictions over the space of labels \citep{yu2020learning}.
When sufficient expert annotation is available, it is preferred to make the most of both worlds.
This means that we would like to use both the expert annotations and at the same time, utilize contrastive techniques in regularizing model consistency for semantically similar examples. 
To do so, we trained the ResNet50 model using the Supervised Contrastive criterion \citep{khosla2020supervised}. The results help shed light on the implicit bias due to the long-tail distribution of labels and the advantages and disadvantages compared to the purely self-supervised approach.

For each image, we apply the probabilistic transformation $\mathcal{T}$ multiple times to obtain $M$ different views. The latent representations are then computed according to Equation \ref{latent_rep}. We will then train a head $g^{\text{supcon}}(\cdot ; \omega)$ which gives us:

\begin{equation}
    u^{(m)}_i = g^{\text{supcon}}(z^{(m)}_i;\omega)\;  \; m \in \{1,2\} \; \wedge \; i \in [N]=\{1, 2, \cdots, N\} \; \; \text{(Projected Representations)}
\end{equation}

In each training batch $\mathcal{B}$ with indices $i_1,..., i_B$, each view $m$ of example $x_i$ will have the following positive set:

\begin{equation}
    P(i) = \{ i_j | y_{i_j}=y_i \}
\end{equation}

This means that in a batch of $B$ items (each item being the projected latent representation of a view of an image), all of the different views of the same image and every view of every image with the same label are considered {\it similar}). The following loss is then used as optimization objective:

\begin{equation}
    \mathcal{L} = - \cfrac{1}{B} \sum_{i=1}^B \cfrac{1}{|P(i)|} \sum_{j \in P(i)} \ln \cfrac{
        \exp((u_i^T \cdot u_j) / \tau)
    }{\sum_{\substack{k \in [B] \\ k \not = i}} \exp((u_i^T \cdot u_k) / \tau)}
\end{equation}

The intuition of this loss is the same as contrastive approaches such as SimCLR and InfoNCE Loss \citep{chen2020simple}; however, the difference is that the {\it positive set} for each example includes all samples with the same label; thus, leveraging the labels while leveraging inter-example similarities to obtain more representative features. Similarly, to perform the final classification head, the projection head $g^{\text{supcon}}(\cdot, \omega)$ is discarded, and a linear classifier is trained on the frozen representations.

\section{Experiments}
\textbf{Dataset:} Here, we utilized \citet{matek2021highly}'s large bone marrow cell image dataset, which is publicly available, to train our pipelines.

The dataset consists of 171374 images (250 x 250 pixels) obtained from 945 patients with a range of reactive and malignant hematological disorders \citep{matek2021highly}. 
All images were obtained at high magnification (40X oil immersion objective) and consisted of a bone marrow cell at the center of the patch, while the image periphery in most cases also contains portions of neighboring marrow cells. The annotations for these images correspond to 21 bone marrow cell types shown in Table  \ref{tab:highly_labels}.

\begin{table}[h!]
\centering
\begin{tabular}{|l|l|l|}
\hline
\textbf{Class}        & \textbf{Abbreviation} & \textbf{Support} \\ \hline
Abnormal eosinophils  & ABE                   & 8                     \\
Artefacts             & ART                   & 19630                 \\
Band Neutrophils      & NGB                   & 9968                  \\
Basophils             & BAS                   & 441                   \\
Blasts                & BLA                   & 11973                 \\
Eosinophils           & EOS                   & 5883                  \\
Erythroblasts         & EBO                   & 27395                 \\
Faggot cells          & FGC                   & 47                    \\
Hairy cells           & HAC                   & 409                   \\
Immature lymphocytes  & LYI                   & 65                    \\
Lymphocytes           & LYT                   & 26242                 \\
Metamyelocytes        & MMZ                   & 3055                  \\
Monocytes             & MON                   & 4040                  \\
Myelocytes            & MYB                   & 6557                  \\
Not identifiable      & NIF                   & 3538                  \\
Other cells           & OTH                   & 294                   \\
Plasma cells          & PLM                   & 7629                  \\
Proerythroblasts      & PEB                   & 2740                  \\
Promyelocytes         & PMO                   & 11994                 \\
Segmented neutrophils & NGS                   & 29424                 \\
Smudge cells          & KSC                   & 42                    \\ \hline            
\end{tabular}
\caption{Dataset statistics from \citet{matek2021highly}\label{tab:highly_labels}}
\end{table}

\textbf{Supervised:}
We first experimented with training the proposed pipeline for a short sequence of only $20$ epochs and cross-entropy as the optimization objective to compare with the baseline performance in \citet{matek2021highly}, the results of which are reported in Table \ref{tab:supall3}. 
The macro-level average of precision and recall over categories were $0.07$ and $0.04$ higher than the reported values in the baseline. The macro average F-1 score of this experiment was $0.07$ higher than the baseline\footnote{
Please note that the baseline F-1 score reported here is computed from the reported values for mean precision and recall, which might be different from the mean F-1 results, as the baseline in \citet{matek2021highly} does not report F-1 values.}, further highlighting the potential of our simple modifications to the training pipeline.

Next, we built a more comprehensive supervised pipeline. We designed two 5-fold experiments with the more extended training sequence of $100$ epochs to ascertain a full convergence. 
Table \ref{tab:sup5foldhorizontal} shows the detailed report of the precision and recall per each class, compared with the state-of-the-art model in \citet{matek2021highly}. 
Our models show consistent improvement across most cell classes over the baseline model. 
Mixup augmentation showed advantages in improving the performance in some of the cell classes; nevertheless, it did not lead to a consistent overall improvement.

\begin{table}[]
\centering
\begin{tabular}{l|ll|}
\cline{2-3}
                                            & \multicolumn{2}{l|}{\textbf{F1-Score}} \\ \hline
\multicolumn{1}{|l|}{\textbf{Class}} & \begin{tabular}[c]{@{}l@{}}Baseline\\ (avg)\end{tabular} & ResNeXt50 \\ \hline
\multicolumn{1}{|l|}{Abnormal eosinophils}  & 0.04          & 0.33                   \\
\multicolumn{1}{|l|}{Artefacts}             & 0.78          & \textbf{0.81}          \\
\multicolumn{1}{|l|}{Basophils}             & 0.23          & \textbf{0.30}          \\
\multicolumn{1}{|l|}{Blasts}                & \textbf{0.70} & \textbf{0.70}          \\
\multicolumn{1}{|l|}{Erythroblasts}         & 0.85          & \textbf{0.91}          \\
\multicolumn{1}{|l|}{Eosinophils}           & 0.88          & \textbf{0.92}          \\
\multicolumn{1}{|l|}{Faggot cells}          & 0.27          & \textbf{0.34}          \\
\multicolumn{1}{|l|}{Hairy cells}           & 0.49          & \textbf{0.52}          \\
\multicolumn{1}{|l|}{Smudge cells}          & 0.43          & \textbf{0.72}          \\
\multicolumn{1}{|l|}{Immature lymphocytes}  & 0.14          & \textbf{0.32}          \\
\multicolumn{1}{|l|}{Lymphocytes}           & 0.79          & \textbf{0.85}          \\
\multicolumn{1}{|l|}{Metamyelocytes}        & 0.41          & \textbf{0.42}          \\
\multicolumn{1}{|l|}{Monocytes}             & \textbf{0.63} & 0.55                   \\
\multicolumn{1}{|l|}{Myelocytes}            & 0.55          & \textbf{0.60}          \\
\multicolumn{1}{|l|}{Band Neutrophils}      & 0.59          & \textbf{0.63}          \\
\multicolumn{1}{|l|}{Segmented neutrophils} & 0.80          & \textbf{0.86}          \\
\multicolumn{1}{|l|}{Not identifiable}      & 0.38          & \textbf{0.51}          \\
\multicolumn{1}{|l|}{Other cells}           & 0.35          & \textbf{0.56}          \\
\multicolumn{1}{|l|}{Proerythroblasts}      & 0.60          & \textbf{0.71}          \\
\multicolumn{1}{|l|}{Plasma cells}          & \textbf{0.82} & \textbf{0.82}          \\
\multicolumn{1}{|l|}{Promyelocytes}         & 0.74          & \textbf{0.75}          \\ \hline
\multicolumn{1}{|l|}{Avg}                   & \textit{0.55} & \textit{\textbf{0.62}} \\ \hline
\end{tabular}
\vspace{1mm}
\caption{ResNeXt-50 trained with cross entropy loss - Short sequence (20 epochs)}
\label{tab:supall3}
\end{table}

\begin{table}[]
\centering
\resizebox{\textwidth}{!}{%
\begin{tabular}{l|llllll|}
\cline{2-7}
 &
  \multicolumn{6}{l|}{\textbf{Precision}} \\ \hline
\multicolumn{1}{|l|}{\textbf{Class}} &
  \textbf{\begin{tabular}[c]{@{}l@{}}Baseline\\ (avg)\end{tabular}} &
  \multicolumn{1}{l|}{\textbf{\begin{tabular}[c]{@{}l@{}}Baseline\\ (std)\end{tabular}}} &
  \textbf{\begin{tabular}[c]{@{}l@{}}ResNext50\\ (avg)\end{tabular}} &
  \multicolumn{1}{l|}{\textbf{\begin{tabular}[c]{@{}l@{}}ResNext50\\ (std)\end{tabular}}} &
  \textbf{\begin{tabular}[c]{@{}l@{}}ResNext50\\ + Mixup (avg)\end{tabular}} &
  \textbf{\begin{tabular}[c]{@{}l@{}}ResNext50\\ + Mixup (std)\end{tabular}} \\ \hline
\multicolumn{1}{|l|}{Abnormal eosinophils} &
  0.02 &
  \multicolumn{1}{l|}{0.03} &
  .4000 &
  \multicolumn{1}{l|}{.5477} &
  .2000 &
  .4472 \\
\multicolumn{1}{|l|}{Artefacts} &
  0.82 &
  \multicolumn{1}{l|}{0.05} &
  .9095 &
  \multicolumn{1}{l|}{.0050} &
  .9086 &
  .0090 \\
\multicolumn{1}{|l|}{Basophils} &
  0.14 &
  \multicolumn{1}{l|}{0.05} &
  .5702 &
  \multicolumn{1}{l|}{.0418} &
  .5349 &
  .0680 \\
\multicolumn{1}{|l|}{Blasts} &
  0.75 &
  \multicolumn{1}{l|}{0.03} &
  .8498 &
  \multicolumn{1}{l|}{.0099} &
  .8181 &
  .0030 \\
\multicolumn{1}{|l|}{Erythroblasts} &
  0.88 &
  \multicolumn{1}{l|}{0.01} &
  .9470 &
  \multicolumn{1}{l|}{.0056} &
  .9492 &
  .0037 \\
\multicolumn{1}{|l|}{Eosinophils} &
  0.85 &
  \multicolumn{1}{l|}{0.05} &
  .9373 &
  \multicolumn{1}{l|}{.0026} &
  .9523 &
  .0056 \\
\multicolumn{1}{|l|}{Faggot cells} &
  0.17 &
  \multicolumn{1}{l|}{0.05} &
  .2147 &
  \multicolumn{1}{l|}{.0844} &
  .2266 &
  .1035 \\
\multicolumn{1}{|l|}{Hairy cells} &
  0.35 &
  \multicolumn{1}{l|}{0.08} &
  .6680 &
  \multicolumn{1}{l|}{.0385} &
  .6320 &
  .0516 \\
\multicolumn{1}{|l|}{Smudge cells} &
  0.28 &
  \multicolumn{1}{l|}{0.09} &
  .6608 &
  \multicolumn{1}{l|}{.1333} &
  .6576 &
  .1317 \\
\multicolumn{1}{|l|}{Immature lymphocytes} &
  0.08 &
  \multicolumn{1}{l|}{0.03} &
  .4058 &
  \multicolumn{1}{l|}{.2265} &
  .3858 &
  .2140 \\
\multicolumn{1}{|l|}{Lymphocytes} &
  0.90 &
  \multicolumn{1}{l|}{0.03} &
  .9029 &
  \multicolumn{1}{l|}{.0021} &
  .9228 &
  .0030 \\
\multicolumn{1}{|l|}{Metamyelocytes} &
  0.30 &
  \multicolumn{1}{l|}{0.05} &
  .4698 &
  \multicolumn{1}{l|}{.0205} &
  .4481 &
  .0057 \\
\multicolumn{1}{|l|}{Monocytes} &
  0.57 &
  \multicolumn{1}{l|}{0.05} &
  .6611 &
  \multicolumn{1}{l|}{.0072} &
  .6789 &
  .0212 \\
\multicolumn{1}{|l|}{Myelocytes} &
  0.52 &
  \multicolumn{1}{l|}{0.05} &
  .6199 &
  \multicolumn{1}{l|}{.0156} &
  .6159 &
  .0110 \\
\multicolumn{1}{|l|}{Band Neutrophils} &
  0.54 &
  \multicolumn{1}{l|}{0.03} &
  .7011 &
  \multicolumn{1}{l|}{.0130} &
  .6837 &
  .0155 \\
\multicolumn{1}{|l|}{Segmented neutrophils} &
  0.92 &
  \multicolumn{1}{l|}{0.02} &
  .9158 &
  \multicolumn{1}{l|}{.0028} &
  .9274 &
  .0065 \\
\multicolumn{1}{|l|}{Not identifiable} &
  0.27 &
  \multicolumn{1}{l|}{0.04} &
  .5155 &
  \multicolumn{1}{l|}{.0216} &
  .5105 &
  .0266 \\
\multicolumn{1}{|l|}{Other cells} &
  0.22 &
  \multicolumn{1}{l|}{0.06} &
  .7748 &
  \multicolumn{1}{l|}{.0466} &
  .8509 &
  .0332 \\
\multicolumn{1}{|l|}{Proerythroblasts} &
  0.57 &
  \multicolumn{1}{l|}{0.09} &
  .6671 &
  \multicolumn{1}{l|}{.0154} &
  .6418 &
  .0239 \\
\multicolumn{1}{|l|}{Plasma cells} &
  0.81 &
  \multicolumn{1}{l|}{0.06} &
  .8838 &
  \multicolumn{1}{l|}{.0129} &
  .8886 &
  .0164 \\
\multicolumn{1}{|l|}{Promyelocytes} &
  0.76 &
  \multicolumn{1}{l|}{0.05} &
  .8883 &
  \multicolumn{1}{l|}{.0107} &
  .8971 &
  .0081 \\ \hline
\multicolumn{1}{|l|}{Avg} &
  \textit{0.51} &
  \multicolumn{1}{l|}{\textit{0.05}} &
  \textit{.6935} &
  \multicolumn{1}{l|}{\textit{.0602}} &
  \textit{.6824} &
  \textit{.0575} \\ \hline
 &
  \multicolumn{6}{l|}{\textbf{Recall}} \\ \hline
\multicolumn{1}{|l|}{\textbf{Class}} &
  \textbf{\begin{tabular}[c]{@{}l@{}}Baseline\\ (avg)\end{tabular}} &
  \multicolumn{1}{l|}{\textbf{\begin{tabular}[c]{@{}l@{}}Baseline\\ (std)\end{tabular}}} &
  \textbf{\begin{tabular}[c]{@{}l@{}}ResNext50\\ (avg)\end{tabular}} &
  \multicolumn{1}{l|}{\textbf{\begin{tabular}[c]{@{}l@{}}ResNext50\\ (std)\end{tabular}}} &
  \textbf{\begin{tabular}[c]{@{}l@{}}ResNext50\\ + Mixup (avg)\end{tabular}} &
  \textbf{\begin{tabular}[c]{@{}l@{}}ResNext50\\ + Mixup (std)\end{tabular}} \\ \hline
\multicolumn{1}{|l|}{Abnormal eosinophils} &
  0.20 &
  \multicolumn{1}{l|}{0.40} &
  .2000 &
  \multicolumn{1}{l|}{.2739} &
  .1000 &
  .2236 \\
\multicolumn{1}{|l|}{Artefacts} &
  0.74 &
  \multicolumn{1}{l|}{0.06} &
  .8312 &
  \multicolumn{1}{l|}{.0025} &
  .8373 &
  .0132 \\
\multicolumn{1}{|l|}{Basophils} &
  0.64 &
  \multicolumn{1}{l|}{0.07} &
  .6077 &
  \multicolumn{1}{l|}{.0277} &
  .6326 &
  .0456 \\
\multicolumn{1}{|l|}{Blasts} &
  0.65 &
  \multicolumn{1}{l|}{0.03} &
  .7934 &
  \multicolumn{1}{l|}{.0178} &
  .8198 &
  .0102 \\
\multicolumn{1}{|l|}{Erythroblasts} &
  0.82 &
  \multicolumn{1}{l|}{0.01} &
  .9232 &
  \multicolumn{1}{l|}{.0033} &
  .9259 &
  .0030 \\
\multicolumn{1}{|l|}{Eosinophils} &
  0.91 &
  \multicolumn{1}{l|}{0.03} &
  .9658 &
  \multicolumn{1}{l|}{.0030} &
  .9636 &
  .0094 \\
\multicolumn{1}{|l|}{Faggot cells} &
  0.63 &
  \multicolumn{1}{l|}{0.27} &
  .2533 &
  \multicolumn{1}{l|}{.1140} &
  .2933 &
  .1437 \\
\multicolumn{1}{|l|}{Hairy cells} &
  0.80 &
  \multicolumn{1}{l|}{0.06} &
  .7922 &
  \multicolumn{1}{l|}{.0382} &
  .7802 &
  .0579 \\
\multicolumn{1}{|l|}{Smudge cells} &
  0.90 &
  \multicolumn{1}{l|}{0.10} &
  .8639 &
  \multicolumn{1}{l|}{.1445} &
  .8861 &
  .1113 \\
\multicolumn{1}{|l|}{Immature lymphocytes} &
  0.53 &
  \multicolumn{1}{l|}{0.15} &
  .2769 &
  \multicolumn{1}{l|}{.1167} &
  .3538 &
  .2586 \\
\multicolumn{1}{|l|}{Lymphocytes} &
  0.70 &
  \multicolumn{1}{l|}{0.03} &
  .8915 &
  \multicolumn{1}{l|}{.0084} &
  .8724 &
  .0029 \\
\multicolumn{1}{|l|}{Metamyelocytes} &
  0.64 &
  \multicolumn{1}{l|}{0.08} &
  .6393 &
  \multicolumn{1}{l|}{.0405} &
  .6822 &
  .0248 \\
\multicolumn{1}{|l|}{Monocytes} &
  0.70 &
  \multicolumn{1}{l|}{0.03} &
  .7856 &
  \multicolumn{1}{l|}{.0160} &
  .7777 &
  .0136 \\
\multicolumn{1}{|l|}{Myelocytes} &
  0.59 &
  \multicolumn{1}{l|}{0.06} &
  .7998 &
  \multicolumn{1}{l|}{.0088} &
  .8048 &
  .0146 \\
\multicolumn{1}{|l|}{Band Neutrophils} &
  0.65 &
  \multicolumn{1}{l|}{0.04} &
  .7232 &
  \multicolumn{1}{l|}{.0046} &
  .7450 &
  .0287 \\
\multicolumn{1}{|l|}{Segmented neutrophils} &
  0.71 &
  \multicolumn{1}{l|}{0.05} &
  .8871 &
  \multicolumn{1}{l|}{.0071} &
  .8748 &
  .0102 \\
\multicolumn{1}{|l|}{Not identifiable} &
  0.63 &
  \multicolumn{1}{l|}{0.04} &
  .6931 &
  \multicolumn{1}{l|}{.0120} &
  .6767 &
  .0090 \\
\multicolumn{1}{|l|}{Other cells} &
  0.84 &
  \multicolumn{1}{l|}{0.06} &
  .8130 &
  \multicolumn{1}{l|}{.0410} &
  .7855 &
  .0390 \\
\multicolumn{1}{|l|}{Proerythroblasts} &
  0.63 &
  \multicolumn{1}{l|}{0.13} &
  .8573 &
  \multicolumn{1}{l|}{.0147} &
  .8734 &
  .0114 \\
\multicolumn{1}{|l|}{Plasma cells} &
  0.84 &
  \multicolumn{1}{l|}{0.04} &
  .9060 &
  \multicolumn{1}{l|}{.0081} &
  .9139 &
  .0167 \\
\multicolumn{1}{|l|}{Promyelocytes} &
  0.72 &
  \multicolumn{1}{l|}{0.08} &
  .7274 &
  \multicolumn{1}{l|}{.0091} &
  .7144 &
  .0224 \\ \hline
\multicolumn{1}{|l|}{Avg} &
  \textit{0.69} &
  \multicolumn{1}{l|}{\textit{0.09}} &
  \textit{.7253} &
  \multicolumn{1}{l|}{\textit{.0434}} &
  \textit{.7292} &
  \textit{.0509} \\ \hline
\end{tabular}%
}
\caption{Supervised setup - ResNeXt-50 - 5-fold experiment trained for 100 epochs}
\label{tab:sup5foldhorizontal}
\end{table}
\textbf{Self-Supervised:} We next turned our attention to a self-supervised learning approach to classify cells in \citet{matek2021highly}'s bone marrow dataset.
One of the significant advantages of using a self-supervised learning approach in this highly class-imbalanced dataset is its robustness to the long-tail distribution of labels, which is known to be a likely cause of performance degradation in imbalanced datasets \citep{jiang2021self}. 


Our configuration for the self-supervised experiment is as follows:
\begin{itemize}
    \item The probabilistic transformation $\mathcal{T}(\cdot)$ that consists of usual transforms, such as color jitter in the HSV color space and random flipping.
    \item $f(\cdot;\Theta)$ is ResNet-50 up until the final pooling layer, returning $z_i \in \mathbb{R}^{2048}$.
    \item $g(\cdot;\omega)$ is a linear layer projecting $z_i$ to a subspace of $\mathbb{R}^{128}$.
    \item The training is carried out for $200$ epochs, with a queue (of size $1280$) being considered starting from epoch $15$ and prototypes $\Omega$ frozen for the first $2$ epochs.
    \item The cardinality of our prototype set is $|\Omega|=1000$. 
    \item The Sinkhorn algorithm is carried out for $3$ iterations, with an epsilon of $\epsilon=0.03$.
    \item The batch size was $256$, and the model was trained on 4 NVIDIA Quadro RTX-5000 GPUs.
\end{itemize}

\begin{table*}[h!]
\resizebox{\textwidth}{!}{%
\begin{tabular}{l|lll|lll|lll|}
\cline{2-10}
 &
  \multicolumn{3}{l|}{\textbf{Precision}} &
  \multicolumn{3}{l|}{\textbf{Recall}} &
  \multicolumn{3}{l|}{\textbf{F1-Score}} \\ \hline
\multicolumn{1}{|l|}{\textbf{Class}} &
  \begin{tabular}[c]{@{}l@{}}Baseline\\ (avg)\end{tabular}&
  \begin{tabular}[c]{@{}l@{}}ResNet50\\ Swav+Lin.\end{tabular} &
  \begin{tabular}[c]{@{}l@{}}ResNet50\\ SupCon\end{tabular} &
  \begin{tabular}[c]{@{}l@{}}Baseline\\ (avg)\end{tabular} &
  \begin{tabular}[c]{@{}l@{}}ResNet50\\ Swav+Lin.\end{tabular} &
  \begin{tabular}[c]{@{}l@{}}ResNet50\\ SupCon\end{tabular} &
  \begin{tabular}[c]{@{}l@{}}Baseline\\ (avg)\end{tabular} &
  \begin{tabular}[c]{@{}l@{}}ResNet50\\ Swav+Lin.\end{tabular} &
  \begin{tabular}[c]{@{}l@{}}ResNet50\\ SupCon\end{tabular} \\ \hline
\multicolumn{1}{|l|}{Abnormal eosinophils} &
  0.02 &
  \textbf{1.00} &
  0.00 &
  0.20 &
  \textbf{1.00} &
  0.00 &
  0.04 &
  \textbf{1.00} &
  0.00 \\
\multicolumn{1}{|l|}{Artefacts} &
  0.82 &
  0.88 &
  \textbf{0.90} &
  0.74 &
  0.89 &
  \textbf{0.90} &
  0.78 &
  0.89 &
  \textbf{0.90} \\
\multicolumn{1}{|l|}{Basophils} &
  0.14 &
  0.72 &
  \textbf{0.77} &
  \textbf{0.64} &
  0.49 &
  0.55 &
  0.23 &
  0.59 &
  \textbf{0.64} \\
\multicolumn{1}{|l|}{Blasts} &
  0.75 &
  0.82 &
  \textbf{0.83} &
  0.65 &
  0.82 &
  \textbf{0.85} &
  0.70 &
  0.82 &
  \textbf{0.84} \\
\multicolumn{1}{|l|}{Erythroblasts} &
  0.88 &
  0.94 &
  \textbf{0.94} &
  0.82 &
  0.95 &
  \textbf{0.95} &
  0.85 &
  0.95 &
  \textbf{0.95} \\
\multicolumn{1}{|l|}{Eosinophils} &
  0.85 &
  \textbf{0.97} &
  0.96 &
  0.91 &
  \textbf{0.95} &
  \textbf{0.95} &
  0.88 &
  \textbf{0.96} &
  0.96 \\
\multicolumn{1}{|l|}{Faggot cells} &
  0.17 &
  0.25 &
  \textbf{0.50} &
  \textbf{0.63} &
  0.10 &
  0.20 &
  0.27 &
  0.14 &
  \textbf{0.29} \\
\multicolumn{1}{|l|}{Hairy cells} &
  0.35 &
  \textbf{0.87} &
  0.79 &
  \textbf{0.80} &
  0.63 &
  0.55 &
  0.49 &
  \textbf{0.73} &
  0.65 \\
\multicolumn{1}{|l|}{Smudge cells} &
  0.28 &
  \textbf{1.00} &
  0.83 &
  \textbf{0.90} &
  0.67 &
  0.56 &
  0.43 &
  \textbf{0.80} &
  0.67 \\
\multicolumn{1}{|l|}{Immature lymphocytes} &
  0.08 &
  0.43 &
  \textbf{0.50} &
  \textbf{0.53} &
  0.23 &
  0.15 &
  0.14 &
  \textbf{0.30} &
  0.24 \\
\multicolumn{1}{|l|}{Lymphocytes} &
  \textbf{0.90} &
  0.88 &
  0.89 &
  0.70 &
  \textbf{0.92} &
  0.92 &
  0.79 &
  0.90 &
  \textbf{0.91} \\
\multicolumn{1}{|l|}{Metamyelocytes} &
  0.30 &
  0.48 &
  \textbf{0.59} &
  \textbf{0.64} &
  0.45 &
  0.49 &
  0.41 &
  0.47 &
  \textbf{0.53} \\
\multicolumn{1}{|l|}{Monocytes} &
  0.57 &
  0.72 &
  \textbf{0.75} &
  0.70 &
  0.72 &
  \textbf{0.72} &
  0.63 &
  0.72 &
  \textbf{0.73} \\
\multicolumn{1}{|l|}{Myelocytes} &
  0.52 &
  \textbf{0.74} &
  0.72 &
  0.59 &
  0.60 &
  \textbf{0.67} &
  0.55 &
  0.66 &
  \textbf{0.69} \\
\multicolumn{1}{|l|}{Band Neutrophils} &
  0.54 &
  0.71 &
  \textbf{0.75} &
  0.65 &
  0.68 &
  \textbf{0.73} &
  0.59 &
  0.70 &
  \textbf{0.74} \\
\multicolumn{1}{|l|}{Segmented neutrophils} &
  0.92 &
  0.89 &
  \textbf{0.90} &
  0.71 &
  0.92 &
  \textbf{0.92} &
  0.80 &
  0.91 &
  \textbf{0.91} \\
\multicolumn{1}{|l|}{Not identifiable} &
  0.27 &
  0.65 &
  \textbf{0.66} &
  \textbf{0.63} &
  0.53 &
  0.56 &
  0.38 &
  0.59 &
  \textbf{0.60} \\
\multicolumn{1}{|l|}{Other cells} &
  0.22 &
  0.85 &
  \textbf{0.88} &
  \textbf{0.84} &
  0.68 &
  0.64 &
  0.35 &
  \textbf{0.75} &
  0.75 \\
\multicolumn{1}{|l|}{Proerythroblasts} &
  0.57 &
  0.79 &
  \textbf{0.80} &
  0.63 &
  \textbf{0.73} &
  0.73 &
  0.60 &
  0.76 &
  0.76 \\
\multicolumn{1}{|l|}{Plasma cells} &
  0.81 &
  0.91 &
  \textbf{0.92} &
  0.84 &
  0.90 &
  \textbf{0.92} &
  0.82 &
  0.91 &
  \textbf{0.92} \\
\multicolumn{1}{|l|}{Promyelocytes} &
  0.76 &
  0.80 &
  \textbf{0.81} &
  0.72 &
  \textbf{0.87} &
  0.84 &
  0.74 &
  \textbf{0.83} &
  0.82 \\ \hline
\multicolumn{1}{|l|}{Avg} &
  \textit{0.51} &
  \textit{\textbf{0.78}} &
  \textit{0.75} &
  \textit{0.69} &
  \textit{\textbf{0.70}} &
  \textit{0.66} &
  \textit{0.55} &
  \textit{\textbf{0.73}} &
  \textit{0.69} \\ \hline
\end{tabular}%
}
\vspace{1mm}
\caption{Self-supervision - Results for SwAV (no labels) and Supervised Contrastive (with labels)}
\label{tab:ssl1}
\end{table*}

\textbf{Supervised Contrastive:} Lastly, following our fully unsupervised experiments, we took a hybrid approach of optimizing the supervised contrastive loss to demonstrate that one can leverage the expert annotations while benefiting from contrastive learning's higher level of robustness to dataset challenges. We performed the experiment with the temperature hyperparameter of $\tau=0.2$. At the core, our $f(\cdot;\Theta)$ was a ResNet-50 architecture truncated at the last pooling output. The projection head $g^{\text{supcon}}(\cdot, \omega)$ was mapping to a subspace of $\mathbb{R}^{128}$. The model was trained with the batch size of $256$ on $4$ NVIDIA Quadro RTX-5000 GPUs.
\begin{figure}
    \centering
    \includegraphics[width=0.99\textwidth]{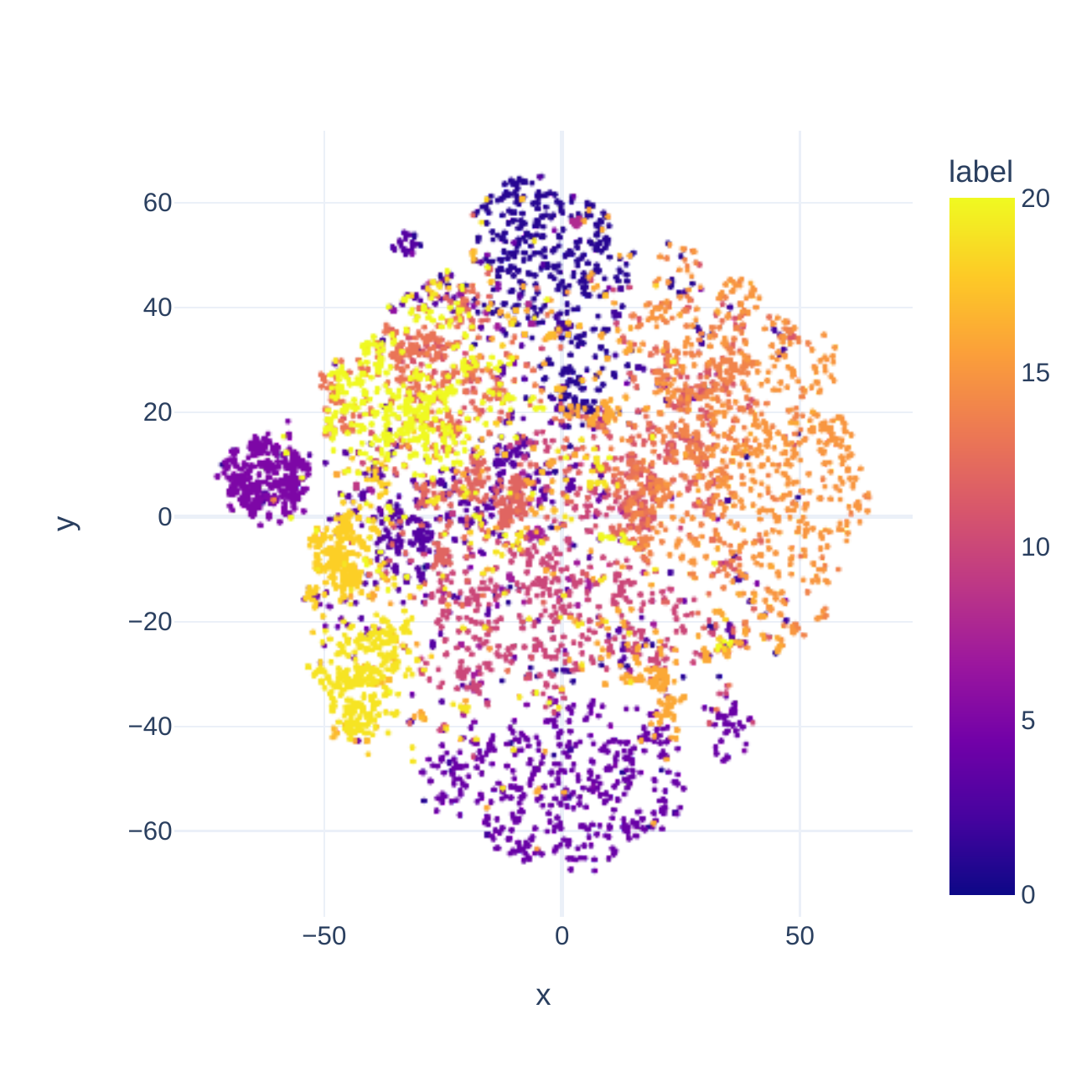}
    \caption{t-SNE visualization of learned unsupervised representations}
    \label{fig:tsne_swav}
\end{figure}

The results of both supervised contrastive and self-supervised methods, shown in Table \ref{tab:ssl1}, demonstrate the advantages of our approach.
Furthermore, the very best results come from the completely unsupervised approach.


\section{Conclusion}
In this work, we proposed solutions based on the principles of self-supervised learning to improve cell classification in the bone marrow cell cytomorphology task. We showed simple modifications to the training pipeline that could lead to a considerable performance boost in the supervised setting. We also prepared a new state-of-the-art baseline for bone marrow cell recognition by leveraging unsupervised visual representation learning, as well as leveraging example similarities in a supervised contrastive setting. Our empirical findings showed considerable improvement in this domain.

\clearpage
\bibliography{iclr2021_conference}
\bibliographystyle{iclr2021_conference}


\end{document}